\RequirePackage{fix-cm}

\documentclass{svjour3}                     
\smartqed  
\usepackage{graphicx}
\usepackage{times}
\usepackage{subfigure} 
\usepackage{amsmath,amssymb,amsfonts}
\usepackage{natbib}

\allowdisplaybreaks

\newcommand{\model}{\mathcal{M}}
\newcommand{\normal}{\mathcal{N}}

\newcommand{\new}{{e+1}}
\newcommand{\expr}{\mathcal{E}}
\newcommand{\data}{\mathcal{D}}

\newcommand{\invwish}{\mathcal{IW}}
\newcommand{\reals}{\mathbb{R}}

\journalname{Machine Learning}
\begin{document}

\title{Retrieval of Experiments with Sequential Dirichlet Process
Mixtures in Model Space
\thanks{This work was financially supported by the Academy of Finland (Finnish Centre of Excellence in Computational Inference
Research COIN, grant no 251170)}
}


\author{Ritabrata Dutta         \and
        Sohan Seth       \and
				Samuel Kaski
}


\institute{Ritabrata Dutta \at
               Department of Information and Computer Science, Aalto University, Finland\\
               \email{ritabrata.dutta@aalto.fi} \\
             \emph{corresponding author}
           \and
           Sohan Seth \at
               Department of Information and Computer Science, Aalto University, Finland\\
               \email{sohan.seth@aalto.fi}
               \and
               Samuel Kaski \at
               Department of Information and Computer Science, Aalto University, Finland\\
               Department of Computer Science, University of Helsinki, Finland\\
               \email{samuel.kaski@aalto.fi} \\
             \emph{corresponding author}                  
}

\date{Received: 06.03.2014 / Accepted: date}

\maketitle


\begin{abstract}
We address the problem of retrieving relevant experiments given a
query experiment,
motivated by the
public databases of datasets in molecular biology and other
experimental sciences, and the need of scientists to relate to earlier
work on the level of actual measurement data.  Since experiments are
inherently noisy and databases ever accumulating, we argue that a
retrieval engine should possess two particular characteristics. First,
it should compare models learnt from the experiments rather than the
raw measurements themselves: this allows incorporating
experiment-specific prior knowledge to suppress noise effects and
focus on what is important. Second, it should be updated sequentially
from newly published experiments, without explicitly storing either
the measurements or the models, which is critical for saving storage 
space and protecting data privacy: this promotes life long learning. We
formulate the retrieval as a ``supermodelling'' problem, of
sequentially learning a model of the set of posterior distributions,
represented as sets of MCMC samples, and suggest the use of 
Particle-Learning-based sequential Dirichlet process mixture (DPM) for this
purpose.  The relevance measure for
retrieval is derived from the \emph{supermodel} through the mixture representation. 
We demonstrate the performance of the proposed retrieval method
on simulated data and molecular biological experiments.

\end{abstract}

\section{Introduction}
\label{intro}
Assume that we have a set of models $\model^e=\{\model_i : i =
1,\ldots,e\}$, each represented by a set of samples from the
posterior distribution given the observed data, that is, $\model_i \equiv \{\theta^{(i)}_j:
j=1,\ldots,n_i\}$. Given an additional query model $\model_\new$, our
task is to find the most relevant models from $\model^e$. This set-up
is motivated by the problem of retrieving relevant experiments given a
query experiment. By an experiment, we imply a set of experimental
data, observations $D_i=\{y_j^{(i)}\}_{j=1}^n$ or
$D_i=\{(x_j^{(i)},y_j^{(i)})\}_{j=1}^n$ where the $x_j^{(i)}$ are
experimental factors or covariates under which the $y_j^{(i)}$ have
been measured. The setting naturally generalizes to any set of data
sets, given suitable models $\model_i$ learned for the $D_i$.

We argue that comparing experiments in the posterior model space has
two distinct advantages. First, it allows incorporating
experiment-specific \emph{prior information} to reduce uncertainty
inherent to any experiment and to indicate what is important in the
data. Second, it allows \emph{not} storing the experiments explicitly
which is critical for saving storage space and protecting data
privacy.

To corroborate our motivation further, consider the example of
biological experiments - a set of measurements over, for instance, a
microarray probe set. Finding relevant existing experiments in this
context is of importance for knowledge transfer, and designing future
experiments. State of the art is comparing manual annotations 
e.g. \citep{malone_modeling_2010}, which
is problematic as the labels are incomplete, the related terminology
varies and develops and, most importantly, annotations are restricted
to known findings. Comparing the actual measurement data would be
desirable but is problematic as well, since the measurements themselves
are inherently noisy, and are usually very few due to price or
availability, in particular in molecular biology.  Learning sensibly
from such measurements requires extensive prior information, for
example in selecting informative variables. Also with the 
accumulation of new datasets from new experiments, storage becomes a bottleneck for
 life long learning where we want to infer sequentially from the new experiments 
 for uncertain future. Hence, our objective is to
build \emph{retrieval engines} that compare experiments in the model
space, and also, to not store either the measurements or the models to
promote life long learning as new measurements become available.
 To the best of our knowledge, this is a novel set-up.

The empirical distribution over the parameter space derived from each 
$\model_i$ represents the corresponding 
posterior distribution given $D_i$. Our strategy is to learn a summary
model or supermodel over the set of posterior distributions
from the previous experiments, to summarize the information in them,
and then exploit this information in retrieval. Hence this supermodel 
will be a latent variable model of the previous posteriors on the 
parameter space. Each of the previous posteriors can then be represented 
as a weighted mixtures of those latent variables with fixed weights for 
each $\model_i$ and the supermodel would contain all the latent variables
and summarize all the previous posteriors jointly. While a new posterior comes, 
we can then represent it using the latent variables of the existing supermodel
and some new latent variables, which will automatically give us the distances between 
the previous posteriors and the new one depending on their shared latent variables and 
their corresponding weights. 

There are two primary challenges in fulfilling this objective. First, modelling the
posterior distributions in the parameter space $\Theta$, and second,
doing it sequentially. In this paper we assume all distributions to be
in the same parameter space, and represented as a set of MCMC samples.
We suggest the use of sequential Dirichlet process mixture model
\citep{zhang_2005,wang_fast_2011} for sequential inference.

Given a large number of samples and a well-behaving model, it is well
known that under some regularity constraints the posterior is approximately
 normally distributed. For small datasets the posterior can be flat, and even multi-modal;
multi-modality can also result from symmetries in the model family. We
argue that in either case a suitable choice of a supermodel is DPM of Gaussian distributions
\citep{ferguson_1973,antoniak_1974}. The consistency of the DPM for
density estimation on parameter space which is Riemannian manifold can be stated from 
\cite{bhattacharya_dunson_2010}. The existing methods using
Gibbs sampling \citep{neal_2000,jain_2004} and variational inference
\citep{blei_2006} to learn DPM models are well explored.  The next
challenge is to learn this mixture model incrementally; that is, while
learning from $\model_j$ we should not need to access
$\model^{j-1}$. To tackle this issue we use a sequential DPM model
\citep{zhang_2005,wang_fast_2011}, with a state-of-the-art particle
learning solution which avoids technical issues in the more
straightforward solutions.
 
DPMs have been extended towards sequential learning before but the
existing algorithms \citep{wang_fast_2011} do not fit our purpose, as
they eventually require access to samples together in order to learn
an appropriate order in which the samples are to be introduced to the
algorithm. We propose a sequential extension that fits our objective
based on particle learning, based on recent works using particle
learning for mixture models in \citep{carvalho_2010,fearnhead_2004,ulker_2010}. 
We exploit this methodology to achieve a true
sequential scheme to explore the parameter space well for multi-modal
density estimation in a high-dimensional space.

We make four contributions in this paper. We (i) introduce the new
retrieval of experiments problem, (ii) propose a solution by
sequential ``supermodelling'' of sets of posterior samples, (iii)
use a state-of-the-art sequential DPM algorithm to solve the
problem, and (iv) demonstrate the retrieval in preliminary experiments
on simulated data and molecular biological data.

To verify our claims, we have showed first by simulation the available
 prior information for different experiments help in improving the performance
  of experiment retrieval in the case of Linear Regression and Sparse Linear
 Regression for High-dimensional dataset. This again supports our intuition
 of using posterior distribution for retrieval of experiments as a tool to
  summarize information from both the prior and data. While showing the performance
   of our sequential experiment retrieval scheme outperforming the performance of
 the best known non-sequential method for different simulated and real life experiments,
  we have covered 4 different kinds of models (namely, Linear Regression, Bayesian Lasso
   for Sparse Linear Regression, Logistic Regression and Factor Models) and used 
   corresponding Bayesian sampling schemes. This shows the adaptability of our method
    for variety of models learnt from different kinds of experiments to retrieve the most relevant experiments.

Our problem set-up resembles multi-task learning and transfer learning
but with some crucial differences. We do not learn directly from the
datasets as in these problems, but from the models. Also, we do not
learn the model of the query dataset using the previously learned
models, and hence do not need to make assumptions about their
relevance \emph{a priori}, we just retrieve them. Our method assumes
that the individual models have been learned before, but that they
have been learned individually to incorporate experiment-specific
knowledge, and not as part of learning a computationally demanding
joint model, which would be the case in multi-task or transfer
learning. That being said, the proposed approach can be seen as
transfer learning in the model space. 

\section{Methodology}
\label{sec:method}
We consider an experiment $\expr$ to be a triplet $\expr\equiv
\{\data, \pi,\model\}$, where $\data$ is a set of $n$ measurements in
$\reals^f$, $\pi$ is the prior knowledge regarding this experiment,
and $\model$ is the posterior of the model, inferred from $\data$ and
$\pi$. To facilitate the inference where a model is learnt from both
$\data$ and $\pi$, we use Bayesian inference. To capture the posterior
distribution well, and to facilitate retrieval in the model space, we
represent each model as a set of MCMC samples in the parameter space $\Theta$.

Given $e$ experiments $\{\expr_1,\expr_2, \ldots, \expr_{e}\}$, our
goal is to summarize the posterior samples by finding the latent
components in the model space and modelling each $\model_i$ as a
mixture of those latent components. Given a new experiment
$\expr_\new$, we decide whether $\model_\new$ can be represented as a
mixture of latent components in the model space similar to the already 
inferred ones or something dissimilar. In either
case this leads to a natural similarity measure between the new
experiment and the existing ones (details below), which can be
exploited for retrieval. A particular model type suitable to model the latent
components of the model space is a Dirichlet Process Mixture (DPM) of Gaussian distributions.
Here, as we consider the models sequentially for faster inference and to prevent the 
cumulative data storage issues, the MCMC samples of the models come in
batches (MCMC samples from one experiment is considered as a batch), hence we need to fit the DPM sequentially.

Dirichlet process mixtures are well known Bayesian nonparametric
models used for consistent density estimation and clustering. They are
special types of product partition models (PPM)
\citep{barry_hartigan_1992}, which induce a probability model on the
partition of the integers $\{1,\ldots,n\}$. Exploiting the connection
between DPMs and PPMs, a sequential learning extension of DPM has
recently been developed \citep{wang_fast_2011} for large
datasets. However, this method depended heavily on the order in which
the observations are presented to the model. The authors optimized the
ordering by calculating a pseudo-marginal likelihood over several
random orderings of the observations, which is obviously not possible
sequentially. In our work we utilize the Particle Learning scheme
developed for Mixture Models by \cite{carvalho_2010}, to implement a
sequential DPM method capable of learning the model over a single
pass. 

In the next two subsections we describe how we do experiment  retrieval with the models
and how to operate the models on the parameter space instead of dataspace. 

\subsection{Retrieval of Experiments}

Here we use sequential DPM scheme on parameter space as the supermodel, 
input data being the MCMC samples coming in batches from each new
model. Given experiments $\{\expr_1,\expr_2, \ldots, \expr_{e}\}$, we model the
 posterior samples
$\{\theta^{(i)}_{j}:i=1,\ldots,e,\,j=1,\ldots,n_e\}$ sequentially in
the parameter space. We make the approximative assumption that they
are simulated from a DP mixture of multivariate normals. This is a
reasonably general assumption although, for instance, discrete-valued
parameters need another model family. So under standard regularity 
assumptions, the corresponding posterior is close to a mixture of Gaussian,
and hence we can use the mixture of Gaussians as supermodel.
We will describe the details of the sequential DPM on parameter space in the next subsection,
but here we will first elaborate on how can we use the supermodel for retrieval of relevant
experiments. 
 

While learning the supermodel, we also simultaneously retrieve relevant experiments
using sequential learning. Supermodel can be used for retrieval of experiments while learning sequentially 
the new query experiment. Given a set of experiments
$\{\expr_1,\ldots,\expr_e\}$ and already learnt supermodel, we sequentially 
learn a DPM from the corresponding MCMC samples
$\{\theta^{(i)}_j:j=1,\ldots,n_i\}_{i=1,\ldots,e}$. Assume the output of DPM from $\{\expr_1,\ldots,\expr_e\}$ is
$\{p_j,\Omega_j\}_{j=1}^h$ where $\{p_j\}$ are the mixing
coefficients, and $\Omega$ are the parameters of the corresponding
mixing distribution $f(\cdot|\Omega)$. Under our assumption of multivariate normal 
assumption, $\Omega=(\mu,\Sigma)$ and $f$ will be the standard multivariate normal density function.
For retrieval, we have already stored the assignment of the MCMC samples to the mixture
components, 
\begin{eqnarray*}
\xi_{j}^{(i)} \equiv \left|\{ \theta^{(i)}_k :
c(\theta^{(i)}_k) = j\}\right|, 
\end{eqnarray*} 
where $c(\theta_k)$ is the component
index of the MCMC sample $\theta_k$. Using $\xi_{j}^{(i)}$ and the mixture 
components we can get back the posteriors for each $\model_i$.

Now given a new query experiment $e+1$, the relevant
experiments to $\expr_{e+1}$ can be found by ranking the earlier
experiments with 
\begin{eqnarray}
\rho(\expr_{e+1}|\expr_l)=\prod_{j=1}^{n_{e+1}} \sum_{k = 1}^h \xi_k^{(l)}
f(\model_j^{(e+1)} | \Omega_k), \,\forall l=1,\ldots,e.
\end{eqnarray}
where $\rho(\expr_{e+1}|\expr_l)$ is the probability of the posterior distribution of
$\expr_{e+1}$ given $\expr_l$ defined by the 
 mixture components $\{\Omega_j\}_{j=1}^h$ and their corresponding
weights $\{\xi_{j=1,\ldots,h}^{(l)}\}_{l=1}^i$. This probability measures how good
a model learnt from the experiment $\expr_l$ would be for  $\expr_{e+1}$ conditioned 
on the supermodel already learnt. To our knowledge this is a 
novel distance measure between two experiments which captures all of the prior 
information and information in datasets, satisfying our criterion of using prior information
without the need of storing datasets.

\subsection{Sequential DPM on Parameter Space}
\label{SDPM}

To implement the sequential DPM algorithm for modelling the 
parameter space, we model each of the latent variables (or
 clusters of MCMC samples) as multivariate normal distribution
with mean $\mu$ and varriance-covariance matrix $\Sigma$. For $\mu, \Sigma$ of each of the 
latent variable or cluster we assume conjugate multivariate-normal-inverse-Wishart
 prior $p_0(\mu,\Sigma)\sim \normal\invwish(\lambda,\kappa,\Sigma,\nu)$ which means
\begin{align*}
\mu|\lambda, \Sigma, \kappa \sim \normal\left(\mu\left|\lambda,\frac{\Sigma}{\kappa}\right.\right),\,
\Sigma|\Omega,\nu\sim\invwish(\Sigma|\Omega,\nu) \; ,
\end{align*}
and the posterior distribution belongs to the same family. 
Using this posterior distribution we can evaluate the likelihood of
a new sample $\theta_{i+1}$ to be assigned to the cluster $c$, given previous
$i$ samples and their cluster allocations ($\gamma_{(i)}$) as a multivariate $t$-distribution:
\begin{equation*}
L_{i+1,c}(\theta_{i+1})= 
\begin{cases}
t_{d_0} \left(\theta_{i+1}|a_0,B_0\Omega\right), \; \; \; c=k_{i}+1 &\\ 
t_{d_{i,c}}
\left(\theta_{i+1}|a_{i,c},B_{i,c}\left[\Omega+.5D_{i,c}\right]\right),
o.w. &
\end{cases}
\end{equation*}
where 
\begin{align*}
 d_0&=2\nu-p+1, a_0=\lambda, B_0={2(\kappa+1)}/{\kappa(2\nu-p+1)},\\
 d_{i,c}&=2 \nu+n_{i,c}-p+1, a_{i,c}={\kappa\lambda+n_{i,c}\bar{\theta}_{i,c}}/{\kappa+n_{i,c}},\\
 B_{i,c}&={2(\kappa+n_{i,c}+1)}/{(\kappa+n_{i,c})(2\nu+n_{i,c}-p+1)},\\
 D_{i,c}&=S_{i,c}+\frac{\kappa n_{i,c}}{\kappa+n_{i,c}}(\lambda-\bar{\theta}_{i,c})(\lambda-\bar{\theta}_{i,c})',\\
 S_{i,c}&=\sum_{\gamma_i=c}(\theta_i-\bar{\theta}_{i,c})(\theta_i-\bar{\theta}_{i,c})',n_{i,c}=6\sum_{j=1}^n\gamma_j=c, \bar{\theta}_{i,c}=\frac{1}{n_{i,c}}\sum_{j:\gamma_j=c}\theta_j.
\end{align*}




To learn DPM of Gaussian we will use Particle learning (PL) scheme. 
The estimate given by PL for DPM is order independent as it shares
information between $N$ particles and each particle has different
assignments. Following PL terminology, we will track particles $Z_i$ specified by
$Z_i=\left\lbrace\gamma^i,\mathbf{s}_i,\mathbf{n}_i,k_i\right\rbrace$,
where the vector $\mathbf{n}_i$ stores the number of observations in
each component and the $\mathbf{s}_i$ contains the sufficient statistics
$\bar{\theta}_{i,c}$ and $S_{i,c}$. In our simulation studies and real life 
examples, we will use $N=100$, and our prior parameters are $\alpha=2$, $\lambda=0$,
$\kappa=0.25$, $\nu=p+2$, and $\Omega= (\nu-0.5(d + 1))I$.

\textbf{Updating Equation for Particle:} When a new observation $\theta_{i+1}$
comes, we will allocate it to the latent variable or cluster
with highest posterior probability and update the parameters of that cluster.
The particle $Z_{i+1}$ is updated as follows:
\begin{itemize}
\item If $\gamma_{i+1}=k_{i}+1$,
\begin{align*} 
m_{i+1}=m_{i}+1,\mbox{ and } s_{i+1,m_{i+1}}=[\theta_{i+1},0]
\end{align*}
\item If $\gamma_{i+1}=c$, 
\begin{align*}
n_{i+1,c}&=n_{i,c}+1, \\
\bar{\theta}_{i+1}&=(n_{i,c}\bar{\theta}_{i,c}+\theta_{i+1})/n_{i+1,c}\\
S_{i+1,c}&=S_{i,c}+\theta_{i+1}\theta_{i+1}'+n_{i,c}\bar{\theta}_{i,c}\bar{\theta}_{i,c}'\\
&\qquad -n_{i+1}\bar{\theta}_{i+1,c}\bar{\theta}_{i+1,c}'.
\end{align*}
\end{itemize}
All other statistics will remain same.

\textbf{Resample the Particles:} In the PL scheme, we will have $N$
particles, $\lbrace{Z^{t}_i\rbrace}_{t=1,\ldots,N}$ and after arrival
of each new observation we will resample $N$ particles from the
existing particles, according to the distribution
$p(\theta_{i+1}|Z_{i}^{t})$ and update these new $N$ particles. Through
this resampling scheme, we share information between $N$ particles
which helps us in achieving the order independence of the
observations.

\textbf{Learning from the Particles:} Each of the particles gives a
mixture normal density estimate on the parameter space. Our final
density estimate or supermodel will be an average over all $N$ particles. 



\newcommand{\bs}[1]{\boldsymbol{#1}}
\renewcommand{\b}[1]{\textbf{#1}}

\section{Simulated Examples}

In this section, first we show the efficacy of the sequential density
estimation scheme on a toy  example explored by \citet{wang_fast_2011} and then
we evaluate the effect of prior information on the retrieval performance. To be
specific, we verify that retrieval performance improves if experiment-specific
prior information is provided.  For our simulated and real life examples, we will
 use a cross validation type set-up where we first learn from all the models sequentially,
 and then for any experiment we retrieve from the rest, rather than doing it on-line.

\paragraph{One-Dimensional Toy Example.}

\begin{figure}[t]
	\centering
		\includegraphics[width=0.45\textwidth]{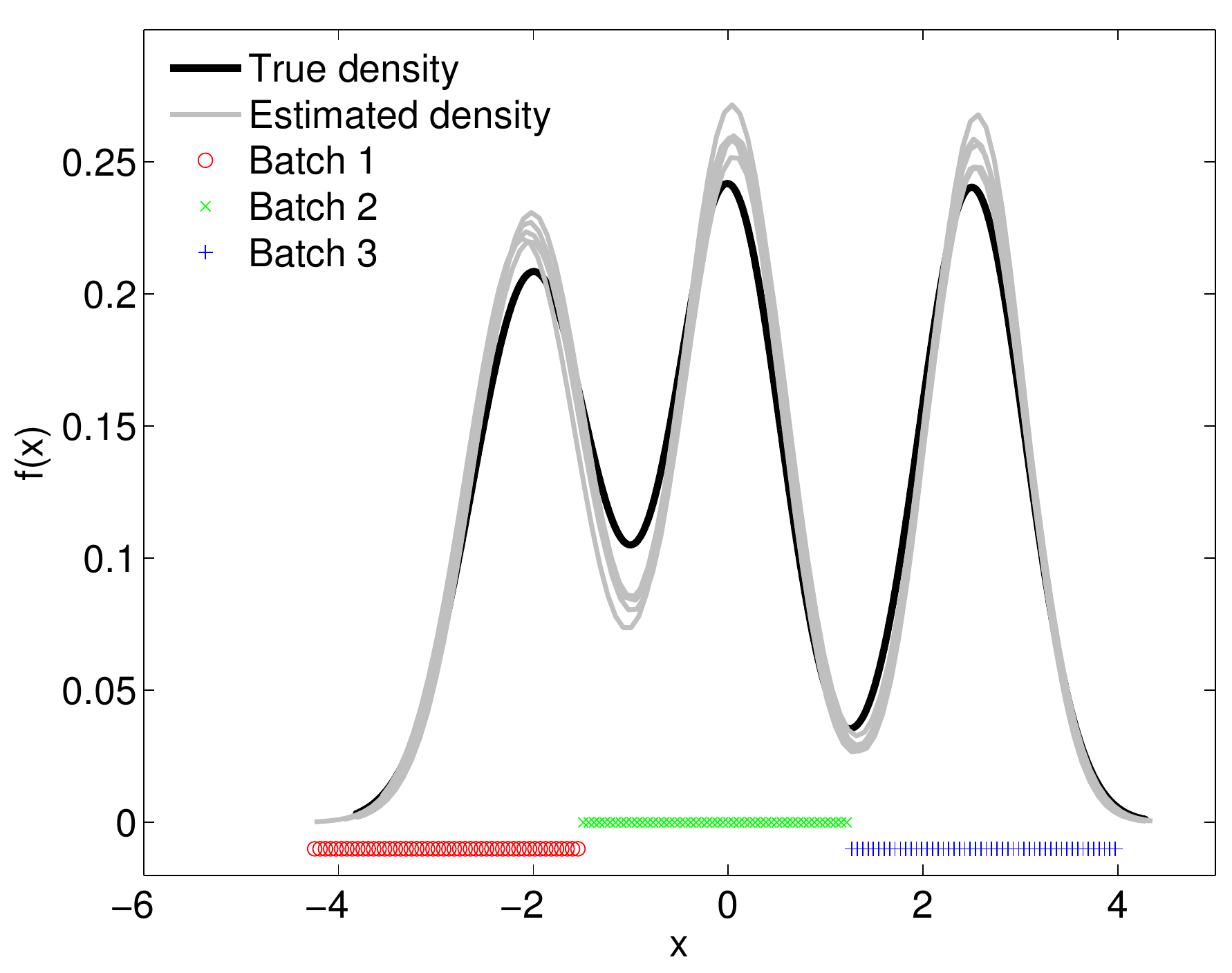}
	\caption{Sequential DPM density estimation from batches of samples. For each of the estimated curves the batches have been presented in random order.}
	\label{MSUGGS-1}
\end{figure}

We consider the 1-dimensional example from \cite{wang_fast_2011}, where the observations are generated from the following density 
\begin{align*}
f(y) = 0.3\normal(y;-2, 0.4)+ 0.5\normal(y; 0, 0.3) + 0.2\normal(y; 2.5, 0.3) \; .
\end{align*}
But as we are interested in learning sequentially, we sample from the three
 modes sequentially and learn the density in a true sequential manner. Thus,
  we have samples $\{y^{(i)}_{j}: j=1,\ldots,n_i,i=1,2,3\}$ where each $i$ 
  corresponds to a particular Gaussian mode. In Figure \ref{MSUGGS-1}, we 
  show the densities estimated by the proposed approach along
with the true density. We observe that the proposed approach is robust in
estimating the true underlying density. Compared to the non-sequential method
in \cite{wang_fast_2011} which optimizes over all possible orderings, the
method considered here can predict the true density in a purely sequential manner.

\paragraph{Effect of Prior Information on Retrieval Performance.} 
We consider two simulated examples of a Bayesian linear regression model,
\begin{displaymath}
y_i=\bs{\beta}^\top \b{x}_i+\epsilon, \mbox{ where }\bs{\beta},\b{x}
\in \reals^p,\mbox{ and }\epsilon\sim\normal(\epsilon|0,\sigma^2) \; ,
\end{displaymath}
with different levels of prior information: In the first example (Case
I), regression problem in low-dimensional, $p=50$, and 
the information in the prior is defined through the distance between the prior
mean and the true model parameters; in the second example (Case II),
we will consider a high-dimensional, $p = 100$, regression problem where
prior information is provided only sparsity assumption (number of
active-inactive components). We compare the performance of the
proposed method (``Sequential") with a non-sequential baseline
(``NSBL") that measures the dissimilarity between two models as the
$L_2$ distance between their respective posterior means. Observe
that for a mean-based model, this baseline will perform very well for
the retrieval purpose. 

We next describe the details on the sampling schemes and on how we
introduce prior information to the models.  We randomly generate
$e=100$ and $e=164$ true models from $10$ and $20$ classes
respectively for the two cases: The true model coefficients were
randomly selected between $[-3,3]$ and $[-2,2]$ respectively, and
noise variance was set to $1$.  For each true model we generate a
random number of observations between $10$-$15$ and $20$-$30$,
respectively for two cases. For both cases we generate $500$ MCMC
samples from each experiment. Prior information will be included
through the parameter $\eta$; its definition is different in the
different cases, and will be explained below.

\textbf{Case I:} We use a normal conjugate prior
$\pi(\bs{\beta}^{(i)}) \sim \normal\left(\bs{\beta}^{(i)}|
\eta\bs{\beta}^{(i)}_0,1\right)$, where $\bs{\beta}^{(i)}_0$ is the
true regressor, and $\eta \in [0,1]$. Therefore, when $\eta=0$, we
essentially provide an uninformative prior whereas when $\eta = 1$ we
provide the true model as the prior. We show the precision-recall
curves as a function of $\eta$ in Figure 2. We can see that for
linear regression our method reaches very close to the optimal level with very little
prior information ($\eta=0.2$).
	
\begin{figure*}[t]
\centering
\subfigure[Case I]{\includegraphics[width=0.48\textwidth]{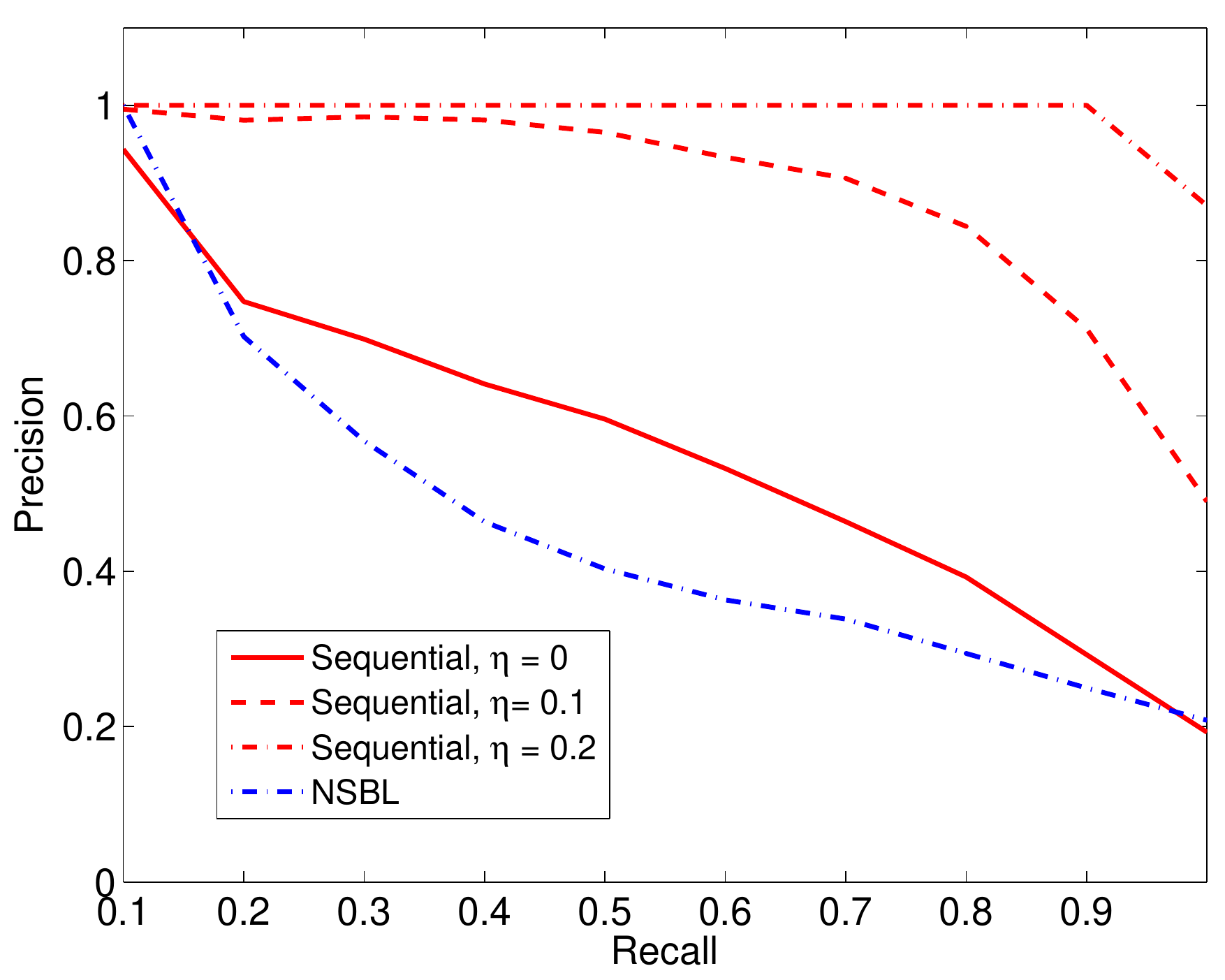}}
\subfigure[Case II]{\includegraphics[width=0.48\textwidth]{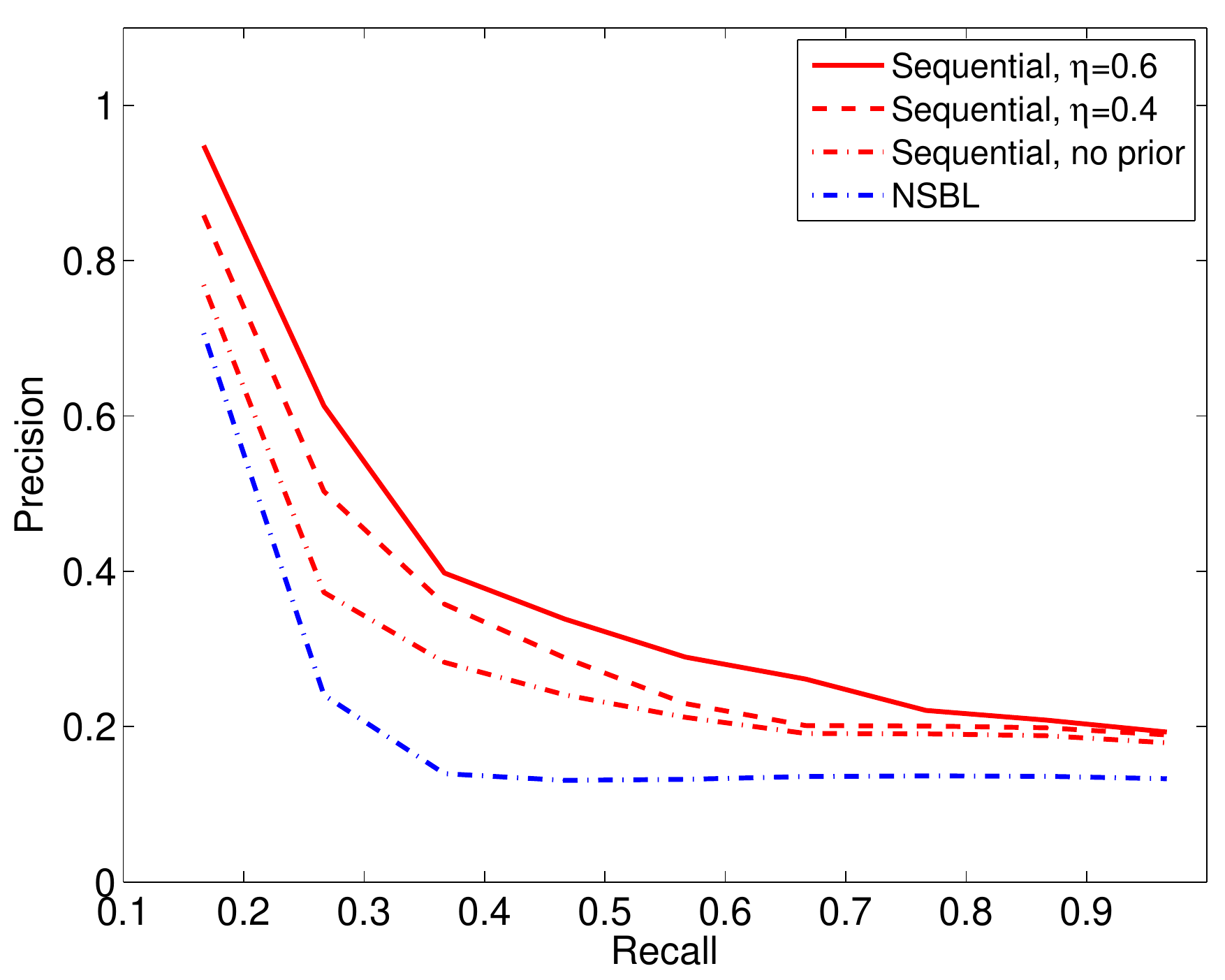}}
\label{fig:prior}
\caption{Effect of varying the amount of prior information on retrieval performance
  in Bayesian linear regression models. ``Sequential" is the proposed
  approach of modelling the parameter space, with $\eta$ describing
  the amount of inserted prior information. The non-sequential
  baseline NSBL retrieves according to the $L_2$ distance between the
  posterior means. It is a proxy for modelling the parameter space
  non-sequentially using standard DPM models, and hence
  computationally impractical due to the dimensionality and ultimate sample
  sizes.}
\end{figure*} 
   
\textbf{Case II:} Following the Bayesian LASSO formulation of \cite{park_casella_2008}, we use the conditional Laplace prior
\begin{align*}
\pi(\bs{\beta}|\sigma^2)=\prod_{j=1}^p{\frac{\lambda_j}{2\sqrt{\sigma^2}}e^{-\lambda_j|\bs{\beta}_j|/2\sqrt{\sigma^2}}}
\end{align*}
and the uninformative scale-invariant marginal prior
$\pi(\sigma^2)=1/\sigma^2$.  Therefore, a higher $\lambda_j$ imposes
sparsity over $\bs{\beta}_j$. We randomly create true models
$\{\bs{\beta}_0^{(i)}\}$ with varying sparse components, say $S_i \in
\{1,\ldots,p\}$. For any true model $\bs{\beta}_0^{(i)}$, we set
$|S_i|$ of the $\lambda_j$'s to $10$ (sparse) and the rest to $1$
(non-sparse). We select $\lambda_j$ at sparse locations 1 (w.p. $\eta$) and 10 (w.p. $1-\eta$) 
and at non-sparse locations 10 (w.p. $\eta$) and 1 (w.p. $1-\eta$). Hence $\eta$ defined the 
amount of true sparsity information in the prior.

For both cases we observe in Figure 2
that the retrieval performance improves consistently as more
experiment-specific prior information is incorporated. In Case II, our
prior information was only regarding the sparsity of the parameters,
not about their real values as in Case I. Hence the improvement due to
to taking the prior into account is smaller. This analysis signifies
the importance of inclusion of true prior knowledge in a Bayesian study, and shown
that the information can be used beneficially in the mentioned of data sets/models.

\section{Experiments}

We validate the performance of the proposed method on one widely used
data set (landmine), and one molecular biology set (human gene
expression); we do not include any additional prior information in
either study. We compare to a non-sequential baseline method that
compares directly posterior means; the expectation is that our method
should perform at the same level or better. The baseline has the
advantage of not having to be sequential, but on the other hand it
only takes the first moments of the distribution into
account. However, the model families are higher-dimensional, and hence
the higher moments are not expected to be very informative. Similar
performances would indicate that the particle learning is able to cope
with the dimensionalities.

\begin{figure*}[t]
\centering
\subfigure[Landmine]{\includegraphics[width=0.48\textwidth]{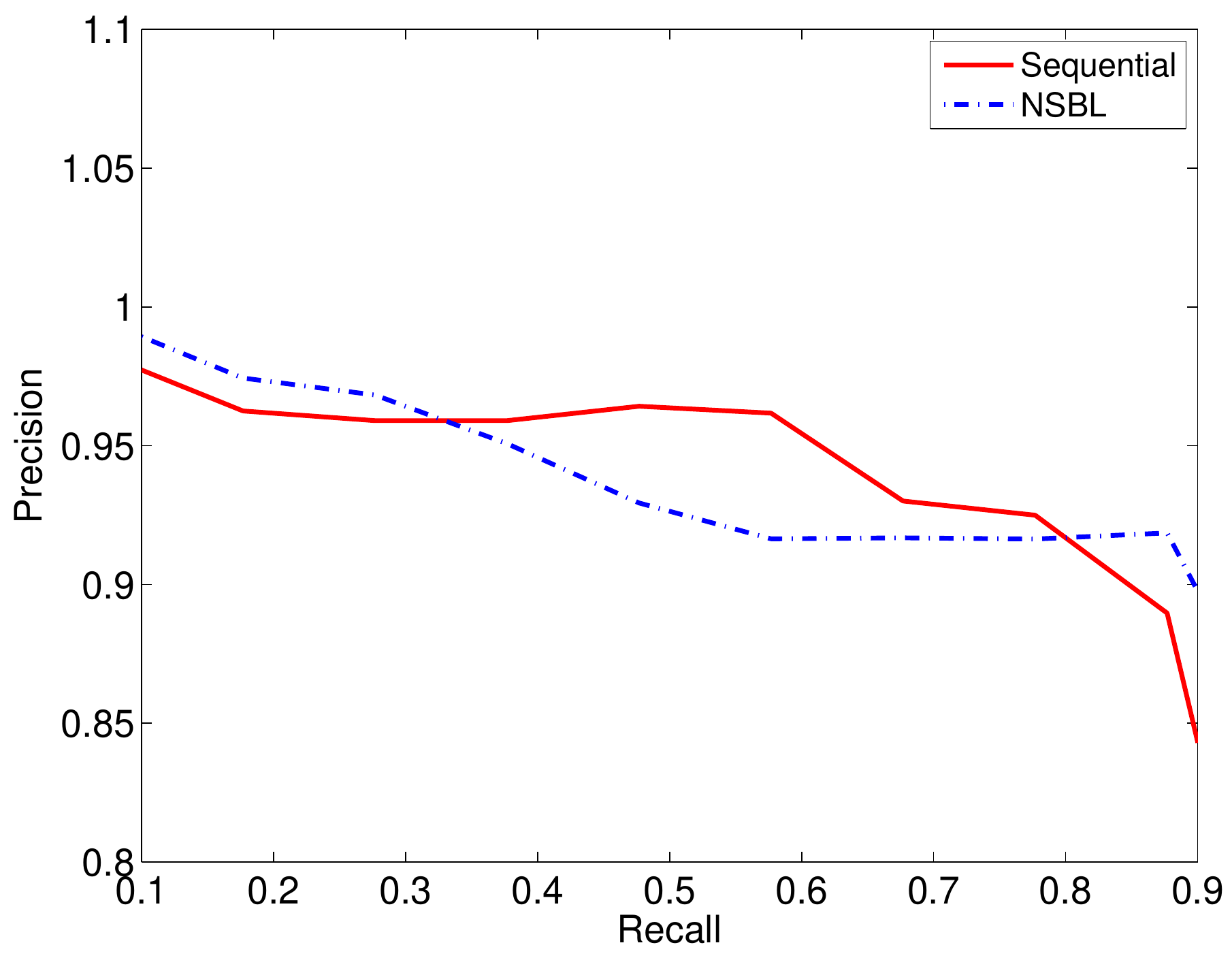}}
\subfigure[Human gene expression]{\includegraphics[width=0.48\textwidth]{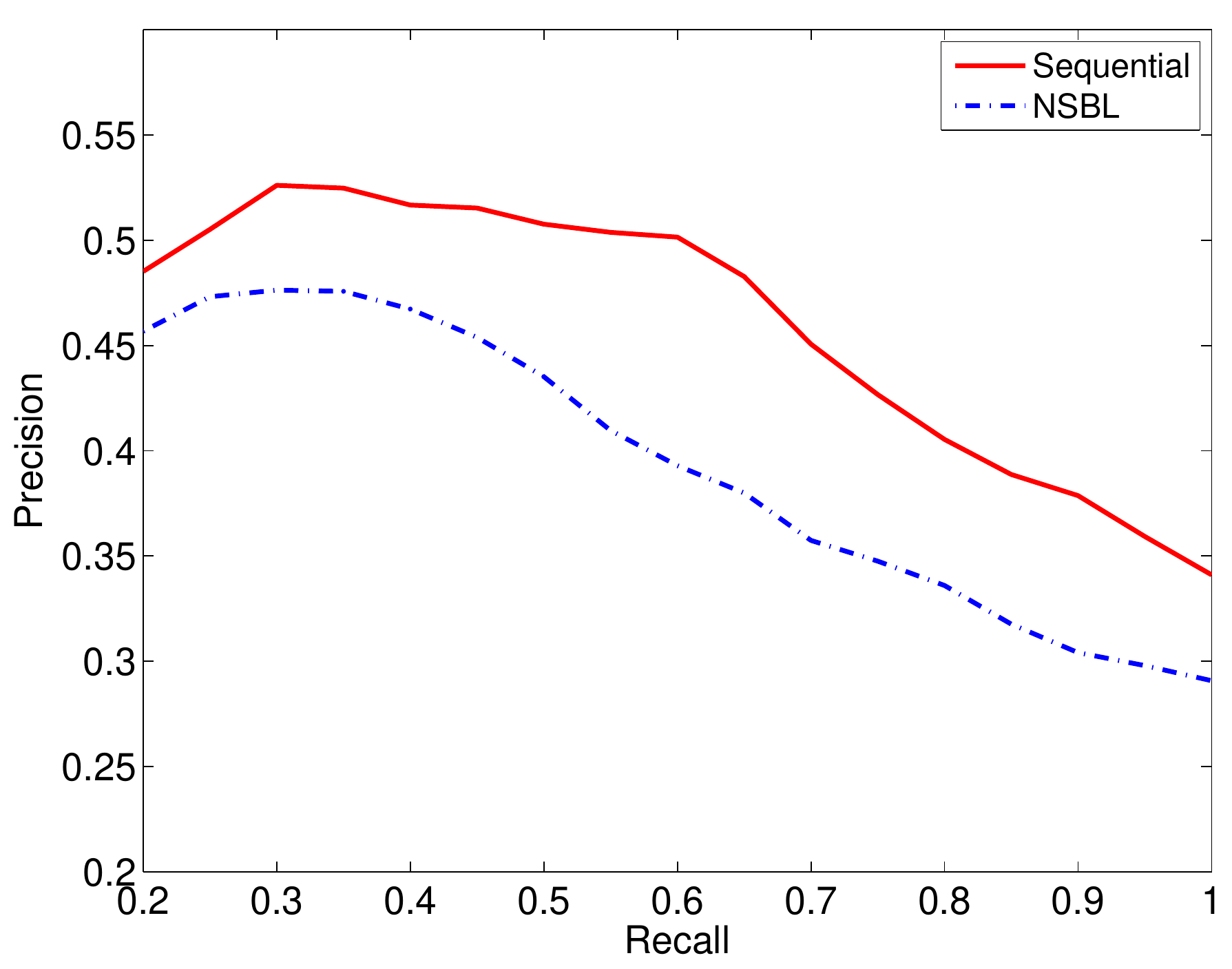}}
\label{fig:real}
\caption{Retrieval performance on real data. ``Sequential" is the
  proposed approach of modelling the parameter space. The non-sequential
  baseline NSBL retrieves according to the $L_2$ distance between the
  posterior means. It is a proxy for modelling the parameter space
  non-sequentially using standard DPM models, and hence
  computationally impractical due to the dimensionality and ultimate sample
  sizes.}
\end{figure*} 
\subsection{Landmine dataset}

We tested the performance of the proposed approach on a landmine
detection dataset \citep{Xue_2007}, consisting of 29 ``experiments.''
Each experiment is a classification task for detecting the presence of
landmine (1) or clutter (0) from 9 input features. Each experiment has
been collected from either a highly foliated region or a desert-like
region. Thus they can be split in two classes (16-13). We train a
Bayesian logistic regression classifier for each experiment with 500
MCMC samples. For MCMC sampling from Bayesian logistic classifier we
have used a block-Gibbs sampler as in \cite{holmes_held_2006}. We
compare the retrieval performance of our algorithm against the same
posterior mean based comparison as in the previous section (Figure 3(a)). We observe
that the sequential method performs as well as the non-sequential
method. This is expected since the two classes are well separable as
reported in the multitask learning literature \citep{Xue_2007}.

\subsection{Human gene expression data}
We consider the human gene expression data from \cite{lukk_2010}. The
data set is a collection of $5372$ absolute 
jointly normalized) gene expression values recorded over
$\sim\!\!22000$ probe sets, mapping to $\sim\!\!14000$ genes. The
collection is divided in $206$ studies, where each study has multiple
annotations, such as cell type, disease state, etc. 
annotation ``15 meta-groups", and We declare the ground truth that two
experiments are similar if they share at least one category in a
particular annotation. We only consider 70 experiments with at least
20 samples, for our evaluation. We pre-process the samples to reduce
the dimensionality of the feature space to 1400 using Gene Set
Enrichment Analysis (GSEA) (applied on probe level for
each sample), and further to 100 by selecting the most informative
gene sets among them.
Hence, we have 70 experiments over a 100 dimensional space. We model each experiment with a factor model, and draw $100$ MCMC samples for each experiment. 
In Figure 3 (b), we compare the performance of the proposed approach with the standard baseline that compares distance between the posterior means. We observe that the non-sequential baseline performs reasonably well in this high-dimensional set-up; however, the sequential retrieval scheme outperforms.

\section{Discussion} 
In this paper we introduce the problem of experiment retrieval and a 
solution where we can use the available prior information and do not need
to store the actual data. The core element of the solution is each experiment
is represented by MCMC samples from the model describing it. The motivation
for using MCMC samples rather than the raw observations is to
incorporate experiment specific prior knowledge as the posterior samples
contain both prior and data information, and we have
demonstrated that this indeed leads to improvement in the retrieval
performance on simulated data. We also were able to get rid of the need of data-storage
by modelling the posterior distributions with an underlying latent variable model.
We introduced the concepts of the supermodel for relating MCMC samples which 
an efficient sequentially computable representation of set of experiments. 
Particle Learning, the sequential scheme for learning
a Dirichlet process mixture under the assumption that the observations
are presented to the algorithm in batches, and each batch has one or
many clusters is used for fast inference. We also solved the issue of data storage creating
the provision for life long learning which suffers from cumulative experiment collections and data 
storage while learning for unlimited time-frame. However, this is a generic set-up that 
can be applied  to other problems as well. 

Though the proposed method performs better in mean-based models like
regression and logistic regression, solutions of very high-dimensional
models, such as factor models of the covariance structure of the
dataset, require further effort. Many model families have
label-switching issues, and the sheer dimensionality is
challenging. We have essentially proposed density estimation in the
model space, and as is well-known, density estimation in ultra-high
dimensionalities is difficult. The current solution is expected to be
useful at least when the parametrization of the model family
significantly compresses the dimensionality of the original data.



\bibliography{ref}
\bibliographystyle{spbasic}

%
%

\end{document}